\documentclass{bmvc2k}
\usepackage{amsmath}
\usepackage{amsthm}
\usepackage{xcolor}

\title{Data-free Parameter Pruning for\\ Deep Neural Networks}

\addauthor{Suraj Srinivas}{surajsrinivas@ssl.serc.iisc.ernet.in}{1}
\addauthor{R. Venkatesh Babu}{venky@serc.iisc.in}{2}

\addinstitution{
 Supercomputer Education and Research Centre,\\
 Indian Institute of Science,\\
 Bangalore, India\\
}

\runninghead{Srinivas, Babu}{Data-free parameter pruning for deep neural networks }

\def\etal{\emph{et al}\bmvaOneDot}

\newtheorem{mydef}{Lemma}
\newtheorem{mydef1}{Result}

\begin{document}

\maketitle

\begin{abstract}
Deep Neural nets (NNs) with millions of parameters are at the heart of many state-of-the-art computer vision systems today. However, recent works have shown that much smaller models can achieve similar levels of performance. In this work, we address the problem of pruning parameters in a trained NN model. Instead of removing individual weights one at a time as done in previous works, we remove one neuron at a time. We show how similar neurons are redundant, and propose a systematic way to remove them. Our experiments in pruning the densely connected layers show that we can remove upto 85\% of the total parameters in an MNIST-trained network, and about 35\% for AlexNet without significantly affecting performance. Our method can be applied on top of most networks with a fully connected layer to give a smaller network.
\end{abstract}

\section{Introduction}
 
\begin{quote}
\textit{I have made this letter longer than usual, only because I have not had the time to make it shorter \footnote{Loosly translated from French}} -~Blaise Pascal
\end{quote}

Aspiring writers are often given the following advice: produce a first draft, then \textit{remove} unnecessary words and \textit{shorten} phrases whenever possible. Can a similar recipe be followed while building deep networks? For large-scale tasks like object classification, the general practice \citep{krizhevsky2012imagenet, Simonyan15, Szegedy_2015_CVPR} has been to use large networks with powerful regularizers \citep{srivastava2014dropout}. This implies that the overall model complexity is much smaller than the number of model parameters. A smaller model has the advantage of being faster to evaluate and easier to store - both of which are crucial for real-time and embedded applications.

Given such a large network, how do we make it smaller? A naive approach would be to remove weights which are close to zero. However, this intuitive idea does not seem to be theoretically well-founded. LeCunn \etal proposed \textit{Optimal Brain Damage}~(OBD) \citep{lecun1989optimal}, a theoretically sound technique which they showed to work better than the naive approach. A few years later, Hassibi \etal came up with \textit{Optimal Brain Surgeon}~(OBS) \citep{hassibi1993second}, which was shown to perform much better than OBD, but was much more computationally intensive. This line of work focusses on pruning unnecessary weights in a trained model. 

There has been another line of work in which a smaller network is trained to mimic a much larger network. Bucila \etal \citep{buciluǎ2006model} proposed a way to achieve the same - and trained smaller models which had accuracies similar to larger networks. Ba and Caruna \citep{ba2014deep} used the approach to show that shallower (but much wider) models can be trained to perform as well as deep models. Knowledge Distillation (KD) \citep{hinton2014distilling} is a more general approach, of which Bucila \etal 's is a special case. FitNets \citep{romero2014fitnets} use KD at several layers to learn networks which are deeper but thinner (in contrast to Ba and Caruna's shallow and wide), and achieve high levels of compression on trained models.
 
Many methods have been proposed to train models that are deep, yet have a lower parameterisation than conventional networks. Collins and Kohli \citep{DBLP:journals/corr/CollinsK14} propose a sparsity inducing regulariser for backpropogation which promotes many weights to have zero magnitude. They achieve reduction in memory consumption when compared to traditionally trained models. Denil \etal \citep{denil2013predicting} demonstrate that most of the parameters of a model can be \textit{predicted} given only a few parameters. At training time, they learn only a few parameters and predict the rest. Ciresan \etal \citep{cirecsan2011high} train networks with random connectivity, and show that they are more computationally efficient than densely connected networks.
 
Some recent works have focussed on using approximations of weight matrices to perform compression. Jenderberg \etal \citep{jaderberg2014speeding} and Denton \etal \citep{denton2014exploiting} use SVD-based low rank approximations of the weight matrix. Gong \etal \citep{gong2014compressing}, on the other hand, use a clustering-based product quantization approach to build an indexing scheme that reduces the space occupied by the matrix on disk. Unlike the methods discussed previously, these do not need any training data to perform compression. However, they change the network structure in a way that prevents operations like fine-tuning to be done easily after compression. One would need to `uncompress' the network, fine-tune and then compress it again.

Similar to the methods discussed in the paragraph above, our pruning method doesn't need any training/validation data to perform compression. Unlike these methods, our method merely prunes parameters, which ensures that the network's overall structure remains same - enabling operations like fine-tuning on the fly. The following section explains this in more detail. 

\section{Wiring similar neurons together}
Given the fact that neural nets have many redundant parameters, how would the weights configure themselves to express such redundancy? In other words, when can weights be removed from a neural network, such that the removal has no effect on the net's accuracy?

Suppose that there are weights which are exactly equal to zero. It is trivial to see that these can be removed from the network without any effect whatsoever. This was the motivation for the naive magnitude-based removal approach discussed earlier.

\begin{figure}[h]
\centering
\includegraphics[width=10cm]{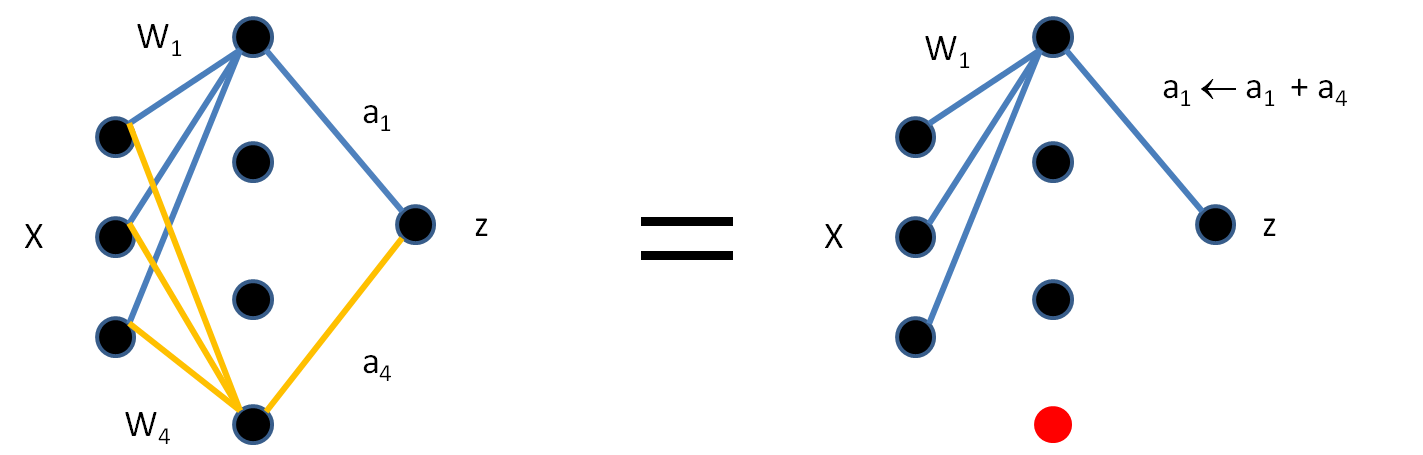}

\caption{A toy example showing the effect of equal weight-sets ($W_1 = W_4$). The circles in the diagram are neurons and the lines represent weights. Weights of the same colour in the input layer constitute a weight-set.}
\label{fig:nn}
\end{figure}

In this work we look at another form of redundancy. Let us consider a toy example of a NN with a single hidden layer, and a single output neuron. This is shown in figure \ref{fig:nn}. Let $W_1, W_2, ... \in \mathcal{R}^d$ be vectors of weights (or `weight-sets') which includes the bias terms, and $a_1, a_2, ... \in \mathcal{R}$ be scalar weights in the next layer. Let $X \in \mathcal{R}^{d}$ denote the input, with the bias term absorbed. The output is given by 

\begin{equation}\label{opn}
z = a_1 h(W_1^T X) + a_2 h(W_2^T X) + a_3 h(W_3^T X) + ... + a_n h(W_n^T X)
\end{equation}  
where $h(\cdot)$ is a monotonically increasing non-linearity, such as sigmoid or ReLU.

Now let us suppose that $W_1 = W_2$. This means that $h(W_1^T X) = h(W_2^T X)$. Replacing $W_2$ by $W_1$ in \eqref{opn}, we get

\begin{equation*}
z = (a_1 + a_2) h(W_1^T X) + 0~h(W_2^T X) + a_3 h(W_3^T X) + ... + a_n h(W_n^T X)
\end{equation*} 

This means whenever two \textit{weight sets} ($W_1,W_2$) are equal, one of them can effectively be removed. Note that we need to alter the co-efficient $a_1$ to $a_1 + a_2$ in order to achieve this. We shall call this the `\textit{surgery}' step. This reduction also resonates with the well-known \textit{Hebbian} principle, which roughly states that ``neurons which fire together, wire together''. If we find neurons that fire together ($W_1 = W_2$), we wire them together ($a_1 = a_1 + a_2$).   Hence we see here that along with single weights being equal to zero, equal weight vectors also contribute to redundancies in a NN. Note that this approach assumes that the same non-linearity $h(\cdot) $ is used for all neurons in a layer.

\section{The case of dissimilar neurons}
Using the intuition presented in the previous section, let us try to formally derive a process to eliminate neurons in a trained network. We note that two weight sets may never be exactly equal in a NN. What do we do when $\|W_1 - W_2\| = \| \epsilon_{1,2} \| \geq 0$ ? Here $\epsilon_{i,j} = W_i - W_j \in \mathcal{R}^d$.

As in the previous example, let $z_n$ be the output neuron when there are $n$ hidden neurons. Let us consider two similar weight sets $W_i$ and $W_j$ in $z_n$ and that we have chosen to remove $W_j$ to give us $z_{n-1}$.
  
We know that the following is true.

\begin{equation*}
z_n = a_1 h(W_1^T X) + ... + a_i h(W_i^T X) + ... + a_j h(W_j^T X) + ...
\end{equation*} 
  
\begin{equation*}
z_{n-1} = a_1 h(W_1^T X) + ... + (a_i + a_j) h(W_i^T X) +  ...
\end{equation*}
 
If $W_i = W_j$ (or $\epsilon_{i,j} = 0$), we would have $z_n = z_{n-1}$. However, since $\| \epsilon_{i,j} \| \geq 0$, this need not hold true. Computing the squared difference $(z_n - z_{n-1})^2$, we have

\begin{equation}\label{mod}
(z_n - z_{n-1})^2 = a_j^2  ( h(W_j^T X) - h(W_i^T X) )^2 
\end{equation} 

To perform further simplification, we use the following Lemma.

\begin{mydef}
Let $a,b \in \mathcal{R}$ and $h(\cdot)$ be a monotonically increasing function, \\such that $max\left( \frac{\mathrm{d}h(x)}{\mathrm{d}x}\right) \leq 1, \forall x \in \mathcal{R}$. Then, 

\begin{equation*}
(h(a) - h(b))^2 \leq (a - b)^2
\end{equation*}

\end{mydef}

The proof for this is provided in the Appendix. Note that non-linearities like sigmoid and ReLU \citep{krizhevsky2012imagenet} satisfy the above property. Using the Lemma and \eqref{mod}, we have 

\begin{equation*}
(z_n - z_{n-1})^2 \leq a_j^2 ~(\epsilon_{i,j}^T X)^2
\end{equation*} 

This can be further simplified using Cauchy-Schwarz inequality.

\begin{equation*}
(z_n - z_{n-1})^2 \leq a_j^2 ~\|\epsilon_{i,j}\|_2^2 ~\|X\|_2^2
\end{equation*}

Now, let us take expectation over the random variable $X$ on both sides. Here, $X$ is assumed to belong to the input distribution represented by the training data.

\begin{equation*}
E (z_n - z_{n-1})^2 \leq a_j^2 ~\|\epsilon_{i,j}\|_2^2 ~ E \|X\|_2^2
\end{equation*}

Note that $E\|X\|_2^2$ is a scalar quantity, independent of the network architecture. Given the above expression, we ask which $(i,j)$ pair least changes the output activation. To answer this, we take minimum over $(i,j)$ on both sides, yielding

\begin{equation}\label{eqfinal}
min (E (z_n - z_{n-1})^2) \leq min(a_j^2 ~\|\epsilon_{i,j}\|_2^2 ) ~  E \|X\|_2^2
\end{equation} 

To minimize an \textit{upper bound} on the expected value of the squared difference, we thus need to find indicies $(i,j)$ such that $a_j^2 ~\|\epsilon_{i,j}\|_2^2$ is the least. Note that we need not compute the value of $E\|X\|_2^2$ to do this - making it dataset independent. Equation \eqref{eqfinal} takes into consideration both the naive approach of removing near-zero weights (based on $a_j^2$) and the approach of removing similar weight sets (based on $\|\epsilon_{i,j}\|_2^2$).

 The above analysis was done for the case of a single output neuron. It can be trivially extended to consider multiple output neurons, giving us the following equation

\begin{equation}
min (E \langle(z_n - z_{n-1})^2\rangle) \leq min(\langle a_j^2\rangle ~\|\epsilon_{i,j}\|_2^2) ~  E \|X\|_2^2
\end{equation} 

where $\langle\cdot \rangle$ denotes the average of the quantity over all output neurons. This enables us to apply this method to intermediate layers in a deep network. For convenience, we define the saliency of two weight-sets in $(i,j)$ as $s_{i,j} = \langle a_j^2\rangle ~\|\epsilon_{i,j}\|_2^2$.

We elucidate our procedure for neuron removal here:
\begin{enumerate}
\item Compute the saliency $s_{i,j}$ for all possible values of $(i,j)$. It can be stored as a square matrix $M$, with dimension equal to the number of neurons in the layer being considered.
\item Pick the minimum entry in the matrix. Let it's indices be $(i',j')$. Delete the $j'^{th}$ neuron, and update $a_{i'} \leftarrow a_{i'} + a_{j'}$. 
\item Update $M$ by removing the $j'^{th}$ column and row, and updating the $i'^{th}$ column (to account for the updated $a_{i'}$.)
\end{enumerate}

The most computationally intensive step in the above algorithm is the computation of the matrix $M$ upfront. Fortunately, this needs to be done only once before the pruning starts, and only single columns are updated at the end of pruning each neuron.

\subsection{Connection to Optimal Brain Damage}
In the case of toy model considered above, with the constraint that only weights from the hidden-to-output connection be pruned, let us analyse the OBD approach.

The OBD approach looks to prune those weights which have the least effect on the training/validation error. In contrast, our approach looks to prune those weights which change the output neuron activations the least. The saliency term in OBD is $s_j = h_{jj} a_j^2/2$, where $h_{ii}$ is the $i^{th}$ diagonal element of the Hessian matrix. The equivalent quantity in our case is the saliency $s_{i,j} = a_j^2 \| \epsilon_{ij} \|_2^2$. Note that both contain $a_j^2$. If the change in training error is proportional to change in output activation, then both methods are equivalent. However, this does not seem to hold in general. Hence it is not always necessary that the two approaches remove the same weights.

In general, OBD removes a single weight at a time, causing it to have a finer control over weight removal than our method, which removes a set of weights at once. However, we perform an additional `surgery' step ($a_i \leftarrow a_i + a_j $) after each removal, which is missing in OBD. Moreover, for large networks which use a lot of training data, computation of the Hessian matrix (required for OBD) is very heavy. Our method provides a way to remove weights quickly.

\subsection{Connection to Knowledge Distillation}

Hinton \etal \citep{hinton2014distilling} proposed to use the `softened' output probabilities of a learned network for training a smaller network. They showed that as $T \rightarrow \infty$, their procedure converges to the case of training using output layer neurons (without softmax).  This reduces to Bucila \etal's \cite{buciluǎ2006model} method. Given a larger network's output neurons $z_l$ and smaller network's neurons $z_s$, they train the smaller network so that $(z_l - z_s)^2$ is minimized.

In our case, $z_l$ corresponds to $z_n$ and $z_s$ to $z_{n-1}$. We minimize an upper bound on \\$E((z_l - z_s)^2)$, whereas KD exactly minimizes $(z_l - z_s)^2$ over the training set. Moreover, in the KD case, the minimization is performed over \textit{all} weights, whereas in our case it is only over the output layer neurons. Note that we have the expectation term (and the upper bound) because our method does not use any training data. 

\subsection{Weight normalization}
In order for our method to work well, we need to ensure that we remove only those weights for which the RHS of \eqref{eqfinal} is small. Let $W_i = \alpha W_j$, where $\alpha$ is a positive constant (say 0.9). Clearly, these two weight sets compute very similar features. However, we may not be able to eliminate this pair because of the difference in magnitudes. We hence propose to normalise all weight sets while computing their similarity.

\begin{mydef1}
For the ReLU non-linearity, defined by $max(0,\cdot)$, and for any $\alpha \in \mathcal{R_+}$ and any $x \in \mathcal{R}$, we have the following result:
\begin{equation*}
max(0,\alpha x) = \alpha~max(0, x)
\end{equation*}
\end{mydef1}

Using this result, we scale all weight sets ($W_1, W_2, ...$) such that their norm is one. The $\alpha$ factor is multiplied with the corresponding co-efficient in the next layer. This helps us identify better weight sets to eliminate. 

\subsection{Some heuristics}
While the mathematics in the previous section gives us a good way of thinking about the algorithm, we observed that certain heuristics can improve performance. 

The usual practice in neural network training is to train the bias without any weight decay regularization. This causes the bias weights to have a much higher magnitude than the non-bias weights. For this reason, we normalize only the non-bias weights. We also make sure that the similarity measure $\epsilon$  takes `sensible-sized' contributions from both weights and biases. This is accomplished for fully connected layers as follows.

Let $W = [W'~b]$, and let $W'_{(n)}$ correspond to the normalized weights. Rather than using $\|\epsilon_{i,j}\| = \|W_i - W_j\|$, we use $\|\epsilon_{i,j}\| = \dfrac{\|W'_{(n)i} - W'_{(n)j}\|}{\|W'_i + W'_j\|} + \dfrac{|b_i - b_j|}{|b_i + b_j|}$.

Note that both are measures of similarity between weight sets. We have empirically found the new similarity measure performs much better than just using differences. We hypothesize that this could be a tighter upper bound on the quantity $E((z_n - z_{n-1})^2)$.

Similar heuristics can be employed for defining a similarity term for convolutional layers. In this work, however, we only consider fully connected layers.

\section{How many neurons to remove?}
One way to use our technique would be to keep removing neurons until the test accuracy starts going below certain levels. However, this is quite laborious to do for large networks with multiple layers. 

\begin{figure}[h]
\begin{tabular}{cc}
\includegraphics[width=6.2cm]{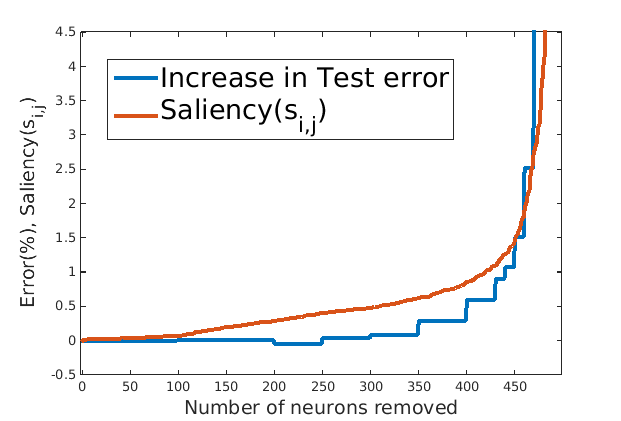}&
\includegraphics[width=6.2cm]{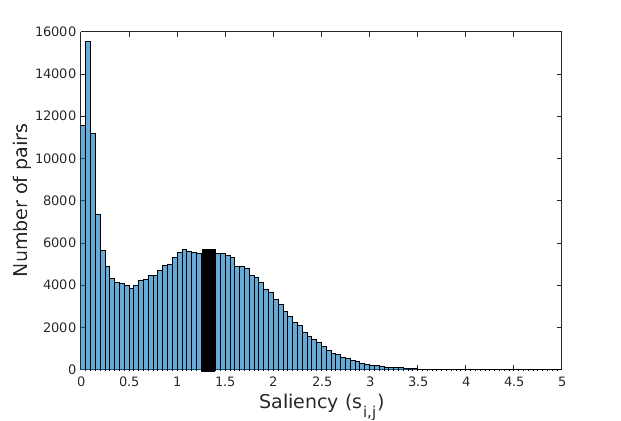}\\
(a)&(b)
\end{tabular}
\caption{(a) Scaled appropriately, the saliency curve closely follows that of increase in test error ; (b) The histogram of saliency values. The black bar indicates the mode of the gaussian-like curve. }
\label{fig:automatic_cutoff}
\end{figure}

We now ask whether it is possible to somehow determine the number of removals automatically. Is there some indication given by removed weights that tell us when it is time to stop? To investigate the same, we plot the saliency $s_{i,j}$ of the removed neuron as a function of the order of removal. For example, the earlier pruned neurons would have a low value of saliency $s_{i,j}$, while the later neurons would have a higher value. The red line in Figure \ref{fig:automatic_cutoff}\textcolor{red}{(a)} shows the same. We observe that most values are very small, and the neurons at the very end have comparatively high values. This takes the shape of a distinct exponential-shaped curve towards the end. 

One heuristic would probably be to have the cutoff point near the foot of the exponential curve. However, is it really justified? To answer the same, we also compute the increase in test error (from baseline levels) at each stage of removal (given by the blue line). We see that the error stays constant for the most part, and starts increasing rapidly near the exponential. Scaled appropriately, the saliency curve could be considered as a proxy for the increase in test error. However, computing the scale factor needs information about the test error curve. Instead, we could use the slope of saliency curve to estimate how densely we need to sample the test error. For example, fewer measurements are needed near the flatter region and more measurements are needed near the exponential region. This would be a \textbf{data-driven} way to determine number of neurons to remove.

We also plot the histogram of values of saliency. We see that the foot of the exponential (saliency $\approx$ 1.2) corresponds to the mode of the gaussian-like curve (Figure \ref{fig:automatic_cutoff}\textcolor{red}{(b)}). If we require a \textbf{data-free} way of finding the number of neurons to remove, we simply find the saliency value of the mode in the histogram and use that as cutoff. Experimentally, we see that this works well when the baseline accuracy is high to begin with. When it is low, we see that using this method causes a substantial decrease in accuracy of the resulting classifier. In this work, we use fractions (0.25,~0.5,~etc) of the number given by the above method for large networks. We choose the best among the different pruned models based on validation data. A truly {data-free} method, however, would require us to not use any validation data to find the number of neurons to prune. Note that only our pruning method is data-free. The formulation of such a complete data-free method for large networks demands further investigation.

\section{Experiments and Results}
In most large scale neural networks \citep{krizhevsky2012imagenet, Simonyan15} , the fully connected layers contain most of the parameters in the network. As a result, reducing just the fully connected layers would considerably compress the network. We hence show experiments with only fully connected layers. 

\subsection{Comparison with OBS and OBD}
Given the fact that Optimal Brain Damage/Surgery methods are very difficult to evaluate for mid-to-large size networks, we attempted to compare it against our method on a toy problem.  We use the SpamBase dataset \citep{asuncion2007uci}, which comprises of 4300 datapoints belonging to two classes, each having 57 dimensional features. We consider a small neural network architecture - with a single hidden layer composed of 20 neurons. The network used a sigmoidal non-linearity (rather than ReLU), and was trained using Stochastic Gradient Descent (SGD). The NNSYSID \footnote{http://www.iau.dtu.dk/research/control/nnsysid.html} package was used to conduct these experiments.

\begin{figure}[h]
\centering
\includegraphics[width=6.2cm]{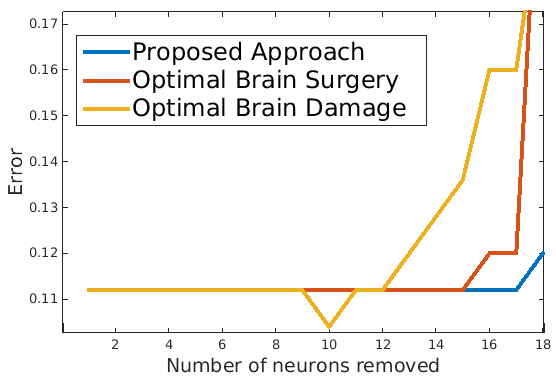}

\caption{Comparison of proposed approach with OBD and OBS. Our method is able to prune many more weights than OBD/OBS at little or no increase in test error }
\label{fig:obd_compare}
\end{figure}

\begin{figure}[h]
\centering
\includegraphics[width=6.8cm]{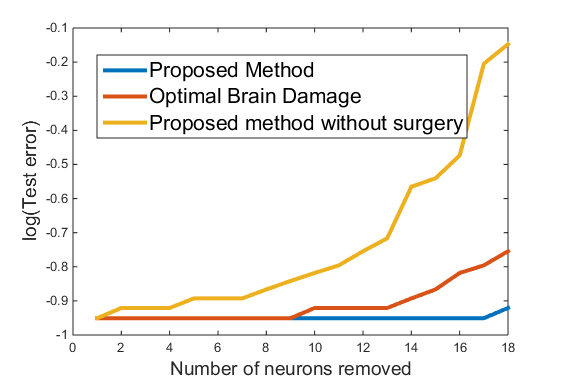}

\caption{ Comparison with and without surgery. Our method breaks down when surgery is not performed. Note that the y-axis is the log of test error.}
\label{fig:obd_compare1}
\end{figure}

Figure \ref{fig:obd_compare} is a plot of the test error as a function of the number of neurons removed. A `flatter' curve indicates better performance, as this means that one can remove more weights for very little increase in test error. We see that our method is able to maintain is low test error as more weights are removed. The presence of an additional `surgery' step in our method improves performance when compared to OBD. Figure \ref{fig:obd_compare1} shows performance of our method when surgery is not performed. We see that the method breaks down completely in such a scenario. OBS performs inferior to our method because it presumably prunes away important weights early on - so that any surgery is not able to recover the original performance level. In addition to this, our method took $<0.1$ seconds to run, whereas OBD took $7$ minutes and OBS took $>5$ hours. This points to the fact that our method could scale well for large networks.

\subsection{Experiments on LeNet}
We evaluate our method on the MNIST dataset, using a LeNet-like \citep{lecun1998gradient} architecture. This set of experiments was performed using the Caffe Deep learning framework \citep{jia2014caffe}. The network consisted of a two $5 \times 5$ convolutional layers with 20 and 50 filters, and two fully connected layers with 500 and 10 (output layer) neurons. Noting the fact that the third layer contains $99\%$ of the total weights, we perform compression only on that layer. 

\begin{table}[h]
\begin{tabular}{|c|c|c|c|c|}
\hline 
Neurons pruned & Naive method & Random removals  & Ours & Compression (\%) \\  
\hline 
150 & 99.05 & 98.63 & 99.09 & 29.81 \\ 
\hline 
300 & 98.63 & 97.81 & 98.98 & 59.62 \\ 
\hline 
400 & 97.60 & 92.07 & 98.47 & 79.54 \\
\hline
\textcolor{red}{420} & 96.50 & 91.37 & \textcolor{red}{98.35} & \textcolor{red}{83.52} \\  
\hline 
\textcolor{red}{440} & 94.32 & 89.25 & \textcolor{red}{97.99} & \textcolor{red}{87.45}\\ 
\hline 
450 & 92.87 & 86.35 & 97.55 & 89.44 \\ 
\hline 
470 & 62.06 & 69.82 & 94.18 & 93.47 \\ 
\hline 
\end{tabular} 
\caption{The numbers represent accuracies in (\%) of the models on a test set.  
`Naive method' refers to removing neurons based on magnitude of weights. The baseline model with 500 neurons had an accuracy of 99.06\%. The highlighted values are those predicted for cutoff by our cut-off selection methods.}
\label{mnist-table}
\end{table}

The results are shown in Table \ref{mnist-table}. We see that our method performs much better than the naive method of removing weights based on magnitude, as well as random removals - both of which are data-free techniques.

Our data-driven cutoff selection method predicts a cut-off of $420$, for a $1\%$ decrease in accuracy. The data-free method, on the other hand, predicts a cut-off of $440$. We see that immediately after that point, the performance starts decreasing rapidly.

\subsection{Experiments on AlexNet}  
For networks like AlexNet \citep{krizhevsky2012imagenet}, we note that there exists two sets of fully connected layers, rather than one. We observe that pruning a given layer changes the weight-sets for the next layer. To incorporate this, we first prune weights in earlier layers before pruning weights in later layers. 

For our experiments, we use an AlexNet-like architecture, called CaffeNet, provided with the Caffe Deep Learning framework. It is very similar to AlexNet, except that the order of max-pooling and normalization have been interchanged. We use the ILSVRC 2012 \citep{ILSVRC15} validation set to compute accuracies in the following table.

\begin{table}[h]
\begin{tabular}{|c|c|c|c|c|}
\hline 
 \# FC6 pruned & \# FC7 pruned  & Accuracy (\%) & Compression (\%) & \# weights removed\\ 
\hline
\textcolor{blue}{2800} & 0 & 48.16 & 61.17 & 37M\\
\hline
2100 & 0 & 53.76 & 45.8 & 27.9M\\
\hline
\textcolor{red}{1400} & 0 & \textcolor{red}{56.08} & 30.57 & 18.6M\\
\hline
\textcolor{red}{700} & 0 & \textcolor{red}{57.68} & 15.28 & 9.3M \\ 
\hline
0 & \textcolor{blue}{2818} & 49.76 & 23.5 & 14.3M\\
\hline
0 & \textcolor{black}{2113} & 54.16 & 17.6 & 10.7M\\
\hline
0 & \textcolor{red}{1409} & \textcolor{red}{56.00} & 11.8 & 7.2M\\
\hline
0 & \textcolor{red}{704}  & \textcolor{red}{57.76} & 5.88 & 3.5M\\
\hline
\textcolor{black}{1400} & \textcolor{blue}{2854} & 44.56 & 47.88 & 29.2M\\
\hline
\textcolor{black}{1400} & \textcolor{black}{2140} & 50.72 & 43.55 & 26.5M\\
\hline
\textcolor{black}{1400} & \textcolor{black}{1427} & 53.92 & 39.22 & 23.9M\\
\hline
\textcolor{red}{1400} & \textcolor{red}{713} & \textcolor{red}{55.6} & 34.89 & 21.27M\\
\hline 
\end{tabular} 

\caption{Compression results for CaffeNet. The first two columns denote the number of neurons pruned in each of the FC6 and FC7 layers. The validation accuracy of the unpruned CaffeNet was found to be 57.84\%. Note that it has 60.9M weights in total. The numbers in \textbf{red} denote the best performing models, and those in \textbf{blue} denote the numbers predicted by our data-free cutoff selection method.}
\label{alexnet-table}
\end{table}

 We observe that using fractions (0.25, 0.5, 0.75) of the prediction made by our data-free method gives us competitive accuracies. We observe that removing as many as 9.3 million parameters in case of 700 removed neurons in FC6 only reduces the base accuracy by 0.2\%. Our best method was able to remove upto 21.3 million weights, reducing the base accuracy by only 2.2\%.

\section{Conclusion}
We proposed a data-free method to perform NN model compression. Our method weakly relates to both Optimal Brain Damage and a form of Knowledge Distillation. By minimizing the expected squared difference of logits we were able to avoid using any training data for model compression. We also observed that the saliency curve has low values in the beginning and exponentially high values towards the end. This fact was used to decide on the number of neurons to prune. Our method can be used on top of most existing model architectures, as long as they contain fully connected layers.  

\section*{Appendix}

\begin{proof}[Proof of Lemma 1]
Given $h(\cdot)$ is monotonically increasing, and $max~\left( \frac{\mathrm{d}h(x)}{\mathrm{d}x}\right) \leq 1,~\forall~x \in \mathcal{R}$. 

\begin{align*}
\implies 0 < \dfrac{dh(x)}{dx} \leq 1
\implies \int_b^a 0 ~\mathrm{d}x < \int_b^a \mathrm{d}h(x) \leq \int_b^a \mathrm{d}x
\implies 0  < h(a) - h(b) \leq a - b
\end{align*}

Since both $h(a) - h(b) > 0$, and $a - b > 0$, we can square both sides of the inequality.

\begin{align*}
(h(a) - h(b))^2 \leq (a - b)^2
\end{align*}

\end{proof}

\section*{Acknowledgement}
We gratefully acknowledge the support of NVIDIA Corporation for the donation of the K40 GPU used for this research.

\bibliography{egbib}
\end{document}